\documentclass[conference]{IEEEtran}

\IEEEoverridecommandlockouts                              

\usepackage{cite}
\usepackage{amsmath,amssymb,amsfonts}
\DeclareMathOperator{\E}{\mathbb{E}}
\usepackage{algorithmic}
\usepackage{graphicx}
\usepackage{textcomp}
\usepackage{xcolor}

\usepackage{graphics} 
\usepackage{booktabs}
\usepackage{float}
\usepackage{subcaption}

\begin{document}

\title{Automated Augmentation with Reinforcement Learning and GANs for Robust Identification of Traffic Signs using Front Camera Images}


\author{Sohini Roy Chowdhury$^{1}$, Lars Tornberg$^{2}$, Robin Halvfordsson$^{3}$, \\ 
Jonatan Nordh$^{3}$, 
Adam Suhren Gustafsson$^{3}$,
Joel Wall$^{3}$, 
Mattias Westerberg$^{3}$, 
Adam Wirehed$^{3}$, \\
Louis Tilloy$^{4}$, 
Zhanying Hu$^{4}$, 
Haoyuan Tan$^{4}$, Meng Pan$^{4}$ 
and Jonas Sj\"oberg$^{3}$
\thanks{$^{1}$ Self Driving, Volvo Cars USA, U. Washington (Affiliate, roych@uw.edu)}%
\thanks{$^{2}$ Machine Learning, AI Center of Excellence, VolvoCars, Sweden}%
\thanks{$^{3}$ Department of Electrical Engineering, Chalmers University of Technology, Gothenburg, Sweden. {\tt\small jonas.sjoberg@chalmers.se}}%
\thanks{$^{4}$ Fung Institute, University of California, Berkeley, California, USA.}%
}

\maketitle


\begin{abstract}
Traffic sign identification using camera images from vehicles plays a critical role in autonomous driving and path planning. However, the front camera images can be distorted due to blurriness, lighting variations and vandalism which can lead to degradation of detection performances. As a solution, machine learning models must be trained with data from multiple domains, and collecting and labeling more data in each new domain is time consuming and expensive. In this work, we present an end-to-end framework to augment traffic sign training data using optimal reinforcement learning policies and a variety of Generative Adversarial Network (GAN) models, that can then be used to train traffic sign detector modules. Our automated augmenter enables learning from transformed nightime, poor lighting, and varying degrees of occlusions using the LISA Traffic Sign and BDD-Nexar dataset. The proposed method enables mapping training data from one domain to another, thereby improving traffic sign detection precision/recall from 0.70/0.66 to 0.83/0.71 for nighttime images.
\end{abstract}

\section{Introduction}
Vehicle path planning and autonomous functionalities rely heavily on timely and robust identification of traffic signs regardless of visibility-related challenges \cite{lit1}. Several works till date \cite{lit1}\cite{lit3} have focused on classification of cropped traffic signs using deep learning models, but there continues to be a lack of end-to-end generalizable methods that treat complete front camera images. Since data augmentation is a key module for robust training of deep learning detectors for traffic signs, we present a novel automated augmenter that can map labelled training data from day to night time domains, while ensuring classification performance enhancement. However, the day and night-time images do not require paired labeling for the model training purposes and can be further extended to weather condition variations as well. Thus, the proposed method can significantly increase the volumes of annotated data for machine learning applications such as robust traffic sign identifications.

Several Generative Adversarial Network (GAN) models have been developed till date for style and textural transformations that can aid daytime to night-time image transformation while preserving features that are crucial for specific region of interest (ROI) classification. One such implementation: deep convolutional GAN (DCGAN) \cite{DCGAN} uses transposed convolutions to generate fake night-time images from a random noise vector followed by a convolutional neural network (CNN) discriminator model that aims at separating real and fake images. Another implementation: super resolution GAN (SRGAN) \cite{SRGAN} uses low resolution night images as generator input and CNNs with residual layers for generator and discriminator design to yield high resolution daytime images. The implementation: styleGAN \cite{StyleGAN} further changes the generator model significantly by a mapping network that uses an intermediate latent space to control style and introduces noise variations at each point in the generator model. Another implementation cycleGAN \cite{cycleGAN}: eliminates the need for paired images for the CNN-based generator model and relies on cyclic consistencies between day to night to day transformations followed by a CNN-based discriminator model. Though all these models produce varying degrees of realistic sceneries and face/car images, they suffer from poor resolution of generated ROIs for traffic sign regions in large field of view images, acquired from automotive grade front cameras. In this work, we implement a bounding box GAN (BBGAN) that minimizes transformations around the ROIs, i.e., traffic sign bounding boxes using a feed-forward generator with U-net \cite{Unet} architecture and a CNN-based discriminator. Additionally, reinforcement learning (RL) models are invoked to identify optimal transformation policies to the traffic sign bounding boxes from a set of 20 policies that involve shear, color transformations and occlusions to generate occlusion-based transformations around traffic signs. Our analysis shows that the optimal RL-based traffic sign modification policies along with the BBGAN generated images collectively generalize day-to-night time images and are robust to vandalism and weather-related occlusions as well. Finally we evaluate the usefulness of the automated augmenter by comparatively analyzing the performance of an object detection system (ODS) without and with the inclusion of augmented data in the ODS training set. 

This paper makes two key contributions. First, a combination of RL \cite{AutoAugment} and BBGAN models \cite{zhang2018} are presented to generate pose and lighting variations along with day-to-night time transformations to traffic sign identification datasets, thereby enabling 4 times data augmentation for a single annotated front camera image. Second, the proposed framework operates on automotive grade wide field camera images to conserve ROI-specific structural and textural information that are significant to traffic sign classification tasks, rather than focusing on cropped traffic sign images only. Also, the manually annotated images for variations in image blurriness, orientation and lighting condition used in this work will enable generalizable bench-marking for new methodologies. The proposed automated augmenter is comparatively evaluated with several image transformation strategies that vary in training complexities to assess its generalizability for data augmentation in ROI-specific classification and detection tasks.

\section{Data Augmentation Methods}
\begin{figure}[ht!]
    \centering
    \includegraphics[width=0.48\textwidth]{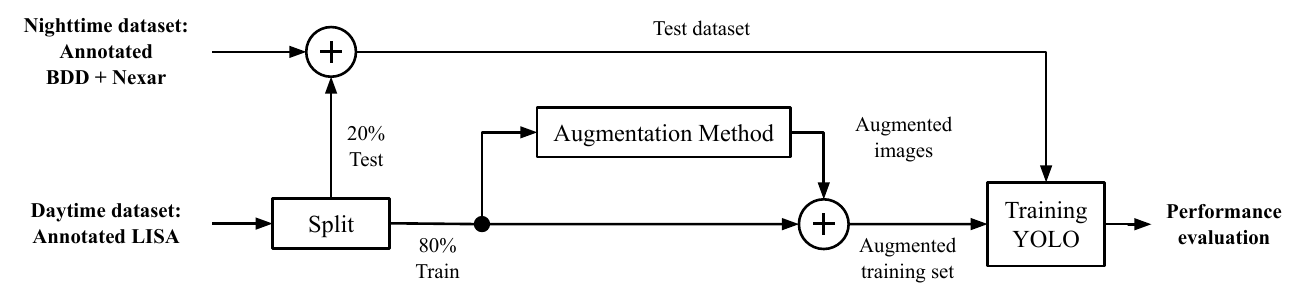}
    \caption{Block diagram of the proposed augmentation method.}
    \label{fig:WOWiterative}
\end{figure}

In this work we improve the ODS performance on data that is out-of-distribution (OOD) with respect to the original training data. The proposed method significantly reduces annotation costs by generating illumination and structural variations to the annotated images, thereby allowing the same annotations to serve multiple images. The architecture of the proposed automated augmenter is shown in Fig. \ref{fig:WOWiterative}. Here, an ODS (YOLOv3 \cite{YOLOv3} network) is trained on daytime images and corresponding artificially transformed nighttime images that are derived through various augmentation methods described below. To establish the performance of the proposed automated augmenter and for baselining purposes, the ODS is trained on 80\% of daytime images from the LISA Traffic Sign Dataset (LISA) \cite{LISA-TS}, and tested on 20\% day time images from the same data set as well as on annotated real night time images from the Berkeley DeepDrive \cite{BerkeleyDeepDrive} and Nexar \cite{Nexar} data sets.

The baseline performance the ODS without data augmentation has precision/recall of 0.897/0.883 and 0.70/0.662 on daytime and night time images, respectively. The discrepancy in the ODS performance between the day and night time images occurs due to the fact that the night time images are OOD. The primary purpose of the automated argumentation methods described below is to increase the performance of the ODS on the night time test data while preserving the detection performance in the day time domain. 

\subsection{Easy Augmentation Methods}
For comparative assessment of automated augmenters with the baseline (no augmentation) ODS performance, we use two easy augmentation methods described below.

\subsubsection{Blender (BLEND)}
The use of 3D-modelling software has previously been used to successfully implement automated pipelines for generation of annotated training data for classifiers \cite{BlenderAug}. Inspired by this we generate traffic signs using Blender \cite{Blender}. We randomly render traffic signs from various angles and backgrounds in collected images of night time traffic scenarios. The world space coordinates of the sign model are automatically transformed to screen space and used as annotations for each rendered image. Examples of BLEND rendered night time images are shown in Fig.\ref{fig:examples_Blender}.

\subsubsection{SimpleAugment (SAUG)}
This method augments the day time images from LISA dataset using three simple pixel-based transformation steps. First, the pixels corresponding to the blue color-plane are decreased based on the initial RGB-vector values per pixel. Here, brighter pixels with higher values are decreased exponentially more than the darker pixels with lower values. This process creates a darker version of the input image. Second, the pixels corresponding to the top half of the image are further decreased in intensity to make the sky region appear darker. Third, the bounding box region corresponding to the traffic signs are retained from the original image to highlight the ROIs. An example of the output of SAUG is shown in Fig. \ref{fig:examples_SAUG}.

\subsection{GAN Models for Domain Transfer and Augmentation} \label{sec:genAugMeth}
\begin{figure}[ht!]
    \centering
    \includegraphics[width=0.3\textwidth]{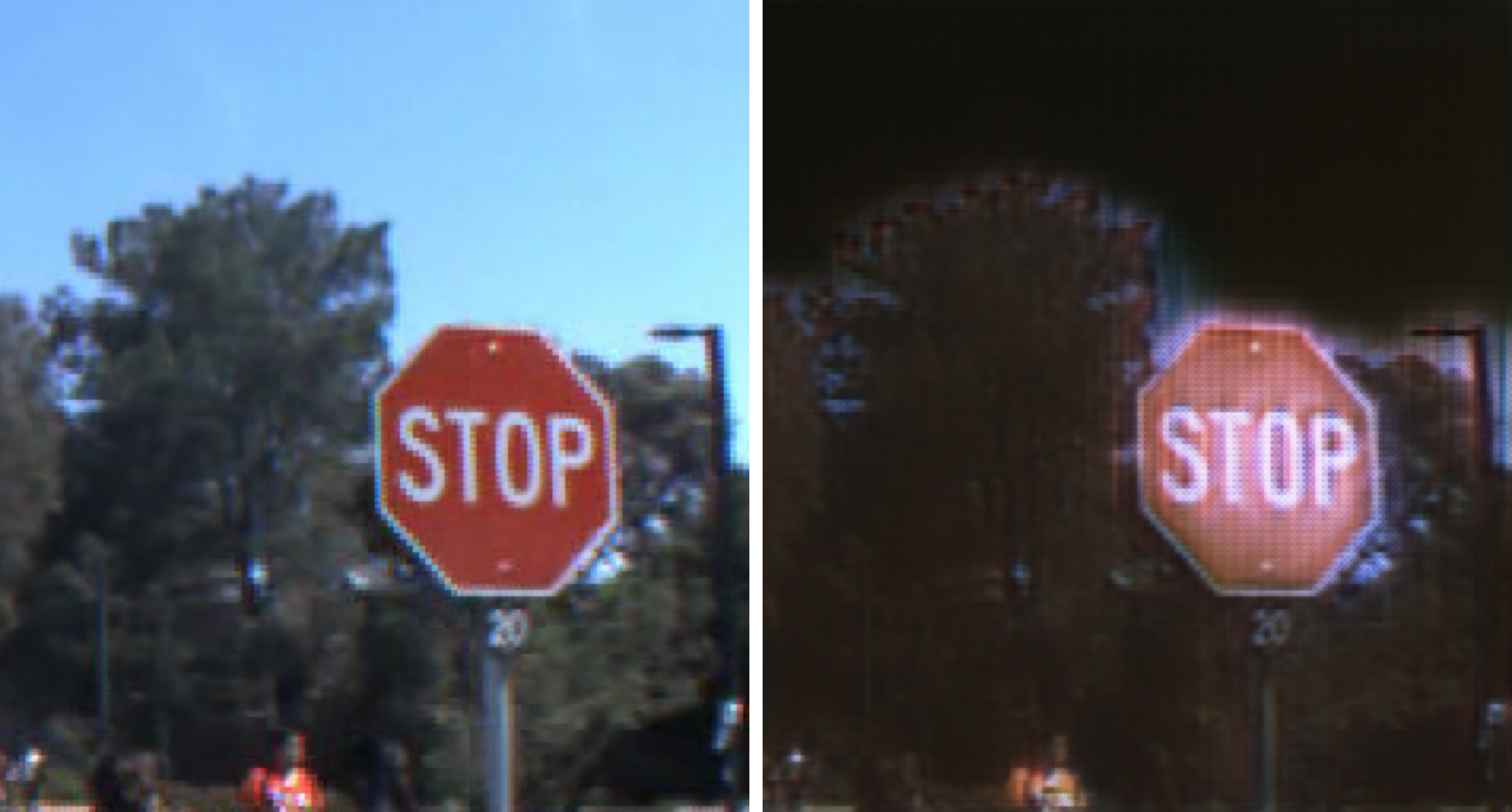}
    \caption{Transformation from day (left) to night (right) images using GAN while maintaining structural definitions for street signs as ROIs.}
    \label{fig:stopSignComp}
\end{figure}

One significant class of methods used for data augmentation are the various variants of GANs. GANs have been used to demonstrate that artificial images can be automatically generated to appear significantly similar to actual camera-acquired or hand-painted images \cite{DCGAN}. This is achieved through a training process that learns an implicit distribution of the training data set $p_x$ from a training set of images $x$. This adversarial training process involves two steps. First step is generation of fake images $z$ following distribution $p_z$ that are minimally dissimilar from real images. This is performed using a trained generator $G$ with parameters $\theta$ that accepts inputs corresponding to image structure and/or image noise. The second step is maximization of discriminatory performance for a classifier $D$ with parameters $\phi$ towards the real and fake images/ROIs in images. The loss ($L$) which is minimized by the GAN optimization routine is given by  
\begin{align}\label{eq1}
L=\min_{\theta}\max_{\phi}\E_{x\sim p_{x}}[log D_{\phi}(x)]+\E_{z \sim p_z}[log(1-D_{\phi}(G_{\theta}(z)))].
\end{align}
The GAN models analyzed in this work and described below are trained using a day time image as input and generating a corresponding night time image as shown in Fig. \ref{fig:stopSignComp}. We observe a daytime image of a stop sign that is converted into its corresponding night time equivalent.
\subsubsection{CycleGAN (CG)}
This method was developed as a tool for domain transfer without the need for paired images from the different domains \cite{cycleGAN}. The CG model, on inference, takes an image as input and outputs the same images with a different style. The key difference from a traditional GAN is that it preserves the content of the input image instead of creating new content from noise. CG comes with the option of choosing different generative models, either a residual network or a U-net architecture. The U-net generates different sections of the image at a time and therefore does not get full context of the image, potentially preventing it from learning certain features. In this work a residual network with significantly high memory requirements is used. Due to this high memory usage, the CG is trained on reduced field of view images (further described in Sec. \ref{sec:datasets}). We observe that CG often generates very dark night-time images, with the corresponding traffic sign being very hard to detect and classify even by a human. To circumvent this problem we implement a version of CG followed by insertions of traffic signs from the daytime image directly. In this way the scene is converted to night time but leaving the content of the ROI unaltered as can be seen in Fig.  \ref{fig:examples_CG_ins}. 
\subsubsection{Bounding Box GAN (BBGAN)}
\begin{figure*}
\begin{subfigure}{0.5\textwidth}
  \centering
  \includegraphics[width=\linewidth]{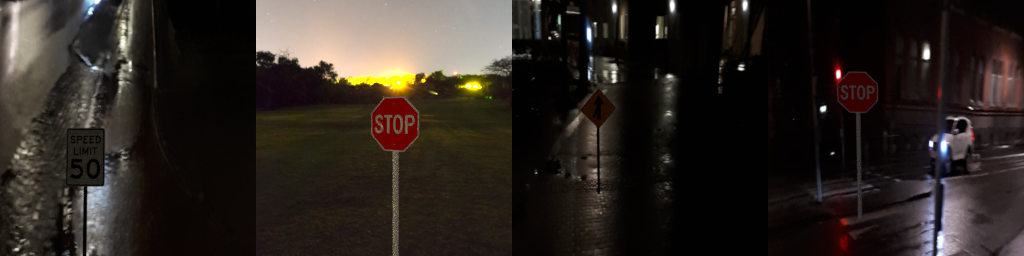}  
  \caption{Example of images generated by BLEND.}
  \label{fig:examples_Blender}
\end{subfigure}
\begin{subfigure}{0.5\textwidth}
  \centering
  \includegraphics[width=\textwidth]{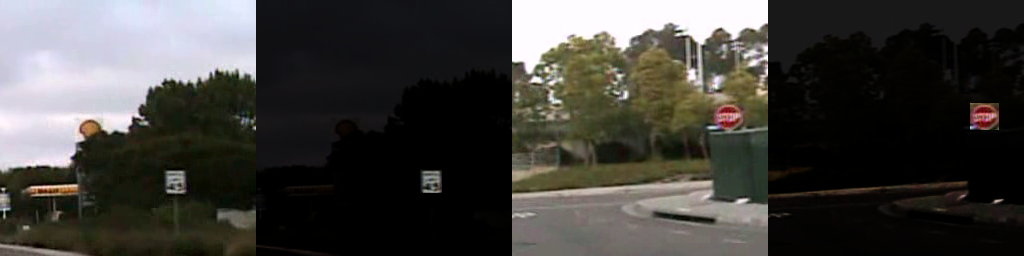}  
  \caption{Example of images generated by SAUG.}
  \label{fig:examples_SAUG}
\end{subfigure}
\newline
\begin{subfigure}{0.5\textwidth}
  \centering
  \includegraphics[width=\linewidth]{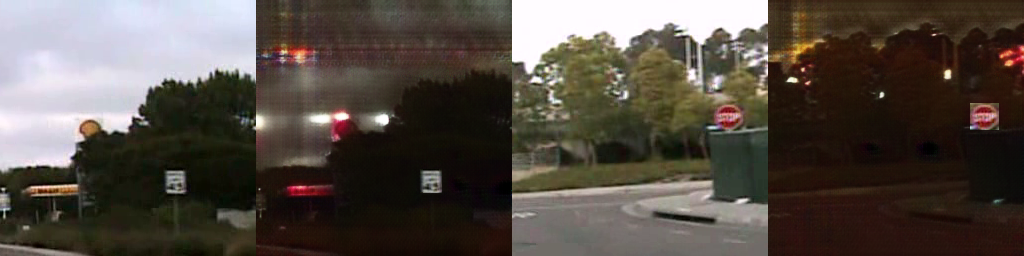}  
  \caption{Example of images generated by CG with traffic signs re-inserted.}
  \label{fig:examples_CG_ins}
\end{subfigure}
\begin{subfigure}{0.5\textwidth}
  \centering
  \includegraphics[width=\linewidth]{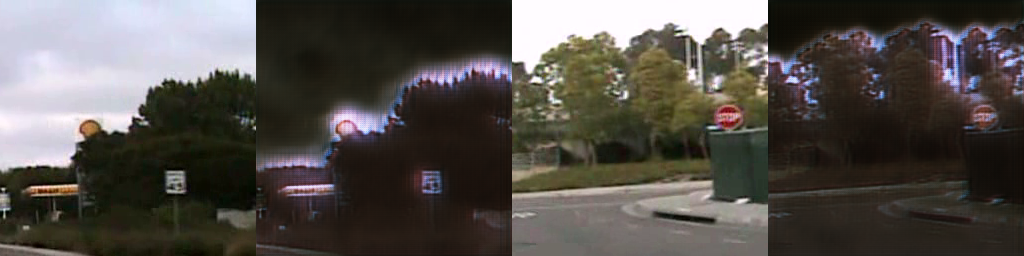}  
  \caption{Example of images generated by BBGAN.}
  \label{fig:examples_BBGAN}
\end{subfigure}
\begin{subfigure}{0.5\textwidth}
  \centering
  \includegraphics[width=\linewidth]{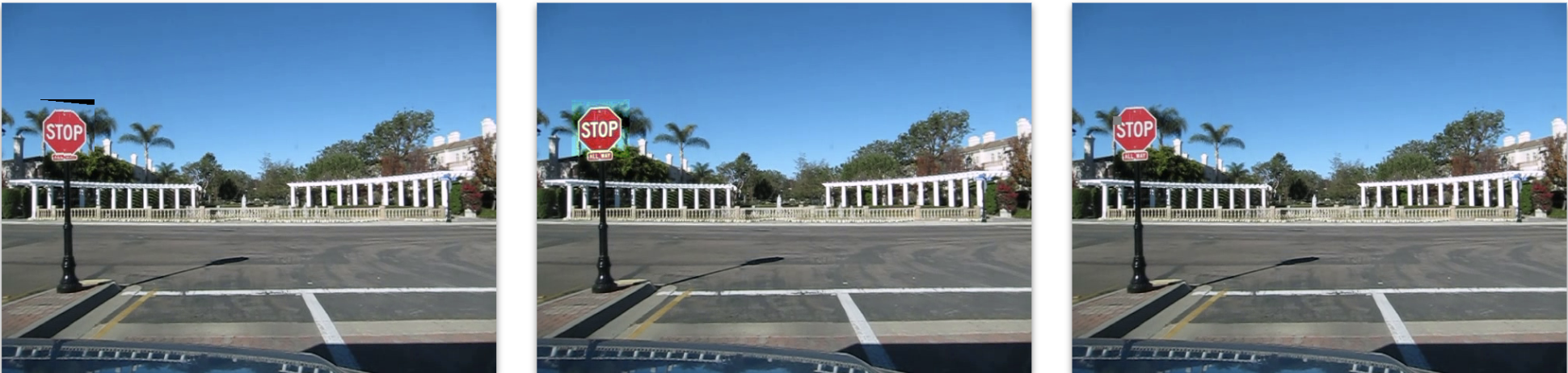}  
  \caption{Example of images generated by RLAUG.}
  \label{fig:examples_AutoAugment}
\end{subfigure}
\begin{subfigure}{0.5\textwidth}
  \centering
  \includegraphics[width=\linewidth]{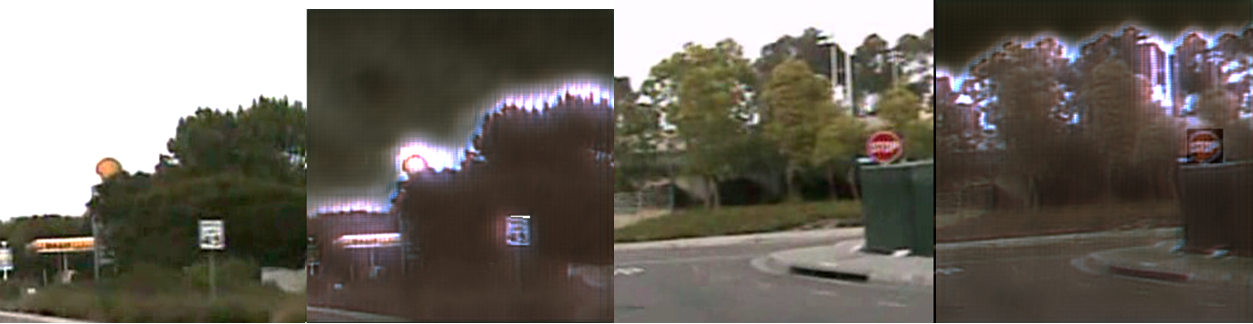}  
  \caption{Examples of augmenting the dataset with BBGAN followed by RLAUG.}
  \label{fig:examples_BBGAN_AutoAugment}
\end{subfigure}
\caption{Examples of automated augmentation using various methods. In subfigures (b) (c) (d) (f), first and third images represent original image, second and fourth image are augmented versions of the preceding image, respectively. }
\label{fig:examples}
\end{figure*}

Although CG is one of the state-of-the art methods for domain transfer, the fact that the resulting images suffered from dark traffic signs limited the performance of the ODS with this augmentation method (as discussed in \ref{sec:datasets}). To preserve the appearance of the traffic signs in the night time images we leverage the fact that we know the location of the traffic sign in the input image. Hence a customized BBGAN, inspired by the work in \cite{L1GAN}, is developed to transfer style from day-to-night time while preserving the content of the bounding box part of the image that contains the traffic sign. The BBGAN minimizes the loss 
\begin{align}\label{eq2}
L=\min_{\theta}\max_{\phi}\E_{x\sim p_{x}}[log D_{\phi}(x)]+\\ \nonumber
\alpha \E_{z \sim p_z}[log(1-D_{\phi}(G_{\theta}(z)))]+\\ \nonumber
(1-\alpha)\E_{x \sim p_x,z\sim p_z}[||\hat{x}-G_{\theta}(\hat{z})||^2_2],    
\end{align}
where, $\alpha$ is a trainable weight parameter and $\hat{x}/\hat{z}$ denote subsets of the images from $x/z$, corresponding to the ROIs.
The last term in Eq.(\ref{eq2}) represents a content preserving loss that penalizes the pixel by pixel difference between input and output image in the ROI. An example of style transfer using BBGAN can be seen in Fig. \ref{fig:examples_BBGAN}. 

\subsection{Reinforcement Learning based Augmentation}
The RL based data augmentation method (RLAUG) in \cite{AutoAugment} automatically searches for image processing policies or operations that can improve ODS performance by data augmentation. This method relies on altering the existing image quality/structure as opposed to the generative networks that directly generate new images as an output. Here, a policy is defined as a sequence of image processing operations to modify an existing image, such as application of rotation, shear and color contrast transformations on images. In \cite{AutoAugment}, a policy consists of 5 sub-policies, and a sub-policy comprises of two operations such that a search algorithm is applied to find the best set of policies that allow an ODS to yield the best validation accuracy on a target dataset. Here, the search algorithm uses a recurrent neural network (RNN) controller, which samples a policy, and a child neural network, which is trained with the policy, to produce a reward signal to update the controller. This augmentation algorithm applies 16 operations such that each operation's probability and magnitude is discretized into uniformly spaced 11 and 10 values, respectively. Thus, the search space for finding 1 policy (containing 5 sub-policies) has about (16 x 10 x 11)$^{10}$ possibilities. The process proceeds as follows: for every image in each batch, one sub-policy is randomly chosen to produce a transformed image to train the child model. In each RNN controller training epoch, the training set is augmented by applying 5 sub-policies to train the child model. The child model is then evaluated to measure the ODS accuracy, which is used as reward signal to train the RNN controller, which in turn gets updated to predict better policies. Thus, the controller samples about 15000 policies for each dataset. Here, a modified policy is created to generate varying degrees of image blurriness and occlusions to learn from vandalized traffic signs as shown in Fig. \ref{fig:examples_AutoAugment}

RLAUG and BBGAN are two standalone automated augmentation methods. However, these methods are combined into one RLAUG$+$BBGAN augmentation method, where the BBGAN first generates night time transformations from the daytime images followed by the RLAUG that converts the day and night time images to their best RL transformed augmented versions. Thus, for each daytime test image in Fig. \ref{fig:examples_BBGAN_AutoAugment}, we obtain its RL transformed version, night time equivalent and RL transformed version of night time image, thus resulting in 4 times data augmentation.
\section{Experiments and Results}\label{sec:datasets}

As discussed in Sec. \ref{sec:genAugMeth}, training GAN-based methods require significant computing power to process large images. The size of images in the LISA dataset varies from [640x480] to [1280x960] pixels. For image standardization all images are cropped to [256x256] pixels while retaining most of the traffic signs withing field of view. The ODS is trained on the LISA training data set containing 7819 images to set a performance baseline on which the different augmentation methods can then be compared. 

The LISA data set contains both the LISA-TS data set and the LISA Extension data set \cite{LISA-TS}. To test the performance on night time images, the trained ODS is tested on 1992 manually annotated real night-time images from the Berkeley DeepDrive \cite{BerkeleyDeepDrive} and Nexar \cite{Nexar} datasets with images containing the same traffic signs as in the test LISA dataset. We have made this annotated data set publicly available\footnote{https://sites.google.com/site/sohiniroychowdhury/research-autonomous-driving} and believe that it makes an important contribution to the development of robust perception systems for autonomous vehicle technology. The traffic sign detector is further tested on 2105 day-time LISA images to assure that the daytime performance remained high despite the additional night-time training data. The data set composition is summarized in Table \ref{tab:datasets}. 

\begin{table}
   \centering
    \caption{Description of the training and testing datasets.}\label{tab:datasets} 
    \scalebox{0.75}
    {
    \begin{tabular}{|c|c|c|c|c|c|}\hline
    {\textbf{Dataset}} & {\textbf{Description}} & {\textbf{Type}} & {\textbf{Use}} & {\textbf{Signs}} & 
        {\textbf{Images}} \\ \hline
        LISA            & LISA TS + LISA Extension (256x256)   & Day   & -     & 10503 & 9924 \\
        LISA test       & LISA test split (~20\%)                       & Day   & Test  & 2161  & 2105 \\
        LISA training   & LISA training split (~80\%)                        & Day   & Train & 8342  & 7819 \\
        BDDNex          & BDD + Nexar (256x256)             & Night & Test  & 2248  & 1992 \\\hline
    \end{tabular}}
    \vspace{-0.5cm}
\end{table}

In the adversarial training process, the mapping between day and night time images requires a training set of night time images, specifically for the discriminator. This training set is extracted from both the Nexar and the BDD datasets. As these data sets contain both night and daytime images, the separation of the night and day time images is included in the pre-processing part of the data pipe-line. The resulting data set for training all GANs consists of 9652 night time images. 
The various augmentation methods are comparatively analyzed for their capability of improving traffic sign detection and classification in terms of precision, which represents the ratio between the number of correctly classified traffic signs and number of positively classified traffic signs per class; and recall, which represents the fraction of actual traffic signs per class that are correctly classified. The training and test data sets are randomly sampled to gauge data sensitivity to the classifier. 

The comparative performance of the ODS without (No Aug) and with various augmentation methods is shown in Table \ref{results}. Here, we observe that the BLEND method yields the best night time traffic sign classification recall. However, it is noteworthy that the BLEND method uses high-quality image textures for each sign as opposed to low-quality real-world examples used by the other methods. This allows the ODS to learn intricate details of each sign and distinguish similar signs. Thus, the BLEND method requires an unscalable amount of manual labor to set up 3D blender models, and may not necessarily generalize well for other types of data augmentation needs.

\begin{table}[!t]
	\centering
	\caption{Performance of YOLOv3 with Augmentation strategies. Mean (std dev).}
	\label{results}
	{
		\begin{tabular}{|l|l|l|l|}\hline
		Augmentation Method & Test Data& Precision & Recall\\ \hline
		&Day&0.897(0.007)&0.883(0.007)\\
		No Aug&Night&0.7(0.01)&0.662(0.01)\\
		&All&0.799(0.006)&0.77(0.007)\\ \hline
		&Day&0.898(0.007)&0.891(0.007)\\
		BLEND&Night&0.788(0.009)&{\bf 0.768(0.009)}\\
		&All&0.842(0.006)&{\bf 0.828(0.006)}\\ \hline
		&Day&0.903(0.007)&0.887(0.007)\\
		SAUG&Night&0.756(0.009)&0.708(0.01)\\
		&All&0.83(0.006)&0.795(0.006)\\ \hline
		&Day&0.91(0.006)&0.893(0.007)\\
		CG&Night&0.716(0.009)&0.712(0.01)\\
		&All&0.811(0.006)&0.8(0.007)\\ \hline
	    &Day&0.907(0.007)&0.895(0.007)\\
		BBGAN&Night&0.761(0.01)&0.677(0.01)\\
		&All&0.836(0.006)&0.783(0.007)\\ \hline
		&Day&0.906(0.006)&0.903(0.006)\\
		RLAUG&Night&0.786(0.009)&0.692(0.01)\\
		&All&0.848(0.006)&0.795(0.006)\\ \hline
	   &Day&{\bf 0.916(0.006)}&{\bf 0.913(0.006)}\\
		\bf{RLAUG$+$BBGAN}&Night&{\bf 0.832(0.008)}&0.707(0.01)\\
		&All&{\bf 0.877(0.005)}&0.808(0.006)\\ \hline
			
	\end{tabular}}
	\vspace{-0.5cm}
\end{table}

With respect to precision, the increase in performance is significant for all methods except for CG. We attribute the low performance of CG to the fact that the content of the traffic signs obtained by this method are too dark to read in many cases. CG produces low quality images because it relies on the ability to map back and forth between the daytime and nighttime domains. The problem is that there is very little information in the night time images which in turn makes a night to day transformation difficult.

The method with best improvement in precision is RLAUG$+$BBGAN where night time recall improved from 0.662 to 0.913. The use of RLAUG to apply policies to the bounding box part of the image does not necessarily provide examples of dark signs. It however allows the ODS to further generalize its identification of well lit signs. This in combination with the ability of BBGAN to increase the night time performance makes it a scalable method that allows for a significant increase in traffic sign classification performance.

\section{Conclusions and Discussion}
\begin{figure}[ht!]
    \centering
    \includegraphics[width=0.5\textwidth, height=4.5in]{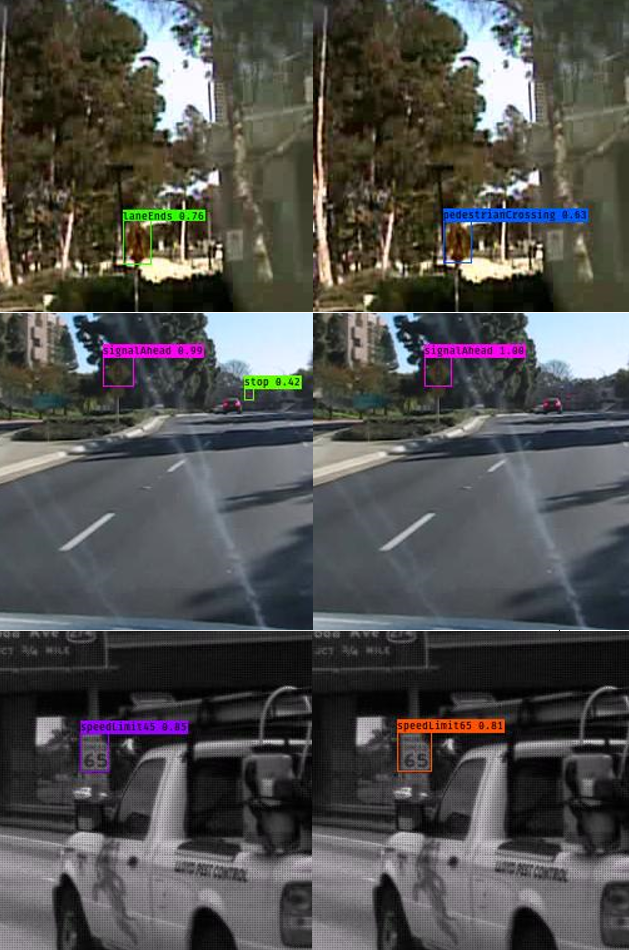}
    \caption{Examples of traffic sign label and respective probabilities. (Left) Yolov3, no augmentation, (Right) Yolov3, RLAUG$+$BBGAN augmenter. }
    \label{fig:correct}
\end{figure}

In this work we demonstrate that for automotive grade front camera images, optimal RL-based modification policies along with GAN generated images collectively generalize day to night time images for traffic sign identification tasks. The proposed automated augmenter aids training object detectors that are robust to image blurriness and vandalism-related occlusions as well. The proposed augmenter (RLAUG$+$BBGAN) enhances precision and recall for traffic sign classification by 3-7\%, while eliminating any test time processing overheads. Examples of improvements in the object detector without and with the proposed augmenter are shown in Fig. \ref{fig:correct}.
Thus, the proposed automated augmenter can be robustly used to train other object detector modules related to autonomous driving and path planning functionalities.

\addtolength{\textheight}{-12cm}   

\bibliographystyle{IEEEtran}
\bibliography{references}

\end{document}